\documentclass[10pt,twocolumn,letterpaper]{article}

\usepackage{iccv}
\usepackage{times}
\usepackage{epsfig}
\usepackage{graphicx}
\usepackage{amsmath}
\usepackage{amssymb}

\usepackage{multirow}
\usepackage[american]{babel}
\usepackage{booktabs}
\usepackage[dvipsnames]{xcolor}
\usepackage{eucal}
\usepackage{bm}

\usepackage[title]{appendix}

\usepackage[breaklinks=true,bookmarks=false]{hyperref}

\usepackage[capitalize]{cleveref}
\crefname{section}{Sec.}{Secs.}
\Crefname{section}{Section}{Sections}
\Crefname{table}{Table}{Tables}
\crefname{table}{Table}{Tables}

\iccvfinalcopy 

\def\iccvPaperID{****} 

\ificcvfinal\fi

\begin{document}

\title{Constraining Depth Map Geometry for Multi-View Stereo: \\ A Dual-Depth Approach with Saddle-shaped Depth Cells}

\author{Xinyi Ye$^{1}$\hspace{0.3in} 
        Weiyue Zhao$^{1}$\hspace{0.3in} 
        Tianqi Liu$^{1}$\hspace{0.3in}
        Zihao Huang$^{1}$\hspace{0.3in}
        Zhiguo Cao$^{1}$\footnotemark[1]\hspace{0.3in}
        Xin Li$^{2}$\hspace{0.1in}\\
$^1${Key Laboratory of Image Processing and Intelligent Control, Ministry of Education; School of Artificial} \\
{ Intelligence and Automation, Huazhong University of Science and Technology, Wuhan 430074, China} \\
$^2$ Department of Computer Science, University of Albany, Albany NY 12222 \\ 
{\tt\small \{xinyiye, zhaoweiyue, tianqiliu, zihaohuang, zgcao\}@hust.edu.cn}\\
{\tt\small xli48@albany.edu}\\
{\small{\url{https://github.com/DIVE128/DMVSNet}}}
\vspace{-2mm}
}

\maketitle
\ificcvfinal\thispagestyle{empty}\fi

\renewcommand{\thefootnote}{\fnsymbol{footnote}} 

\footnotetext[1]{Corresponding author} 

\begin{abstract}
Learning-based multi-view stereo (MVS) methods deal with predicting accurate depth maps to achieve an accurate and complete 3D representation. Despite the excellent performance, existing methods ignore the fact that a suitable depth geometry is also critical in MVS. In this paper, we demonstrate that different depth geometries have significant performance gaps, even using the same depth prediction error. Therefore, we introduce an ideal depth geometry composed of \textbf{Saddle-Shaped Cell}s, whose predicted depth map oscillates upward and downward around the ground-truth surface, rather than maintaining a continuous and smooth depth plane. To achieve it, we develop a coarse-to-fine framework called Dual-MVSNet (DMVSNet),  which can produce an oscillating depth plane. Technically, we predict two depth values for each pixel (\textbf{Dual-Depth}), and propose a novel loss function and a checkerboard-shaped selecting strategy to constrain the predicted depth geometry. Compared to existing methods,DMVSNet achieves a high rank on the DTU benchmark and obtains the top performance on challenging scenes of Tanks and Temples, demonstrating its strong performance and generalization ability. Our method also points to a new research direction for considering depth geometry in MVS.

\end{abstract}

\section{Introduction}

Multi-view stereo~(MVS) is a fundamental technique that bridges the gap between 2-D photograph clues and 3-D spatial information. It takes multiple 2-D RGB observations, as well as their respective camera parameters, to reconstruct the 3-D representation of a scene. There are many applications for MVS that range from autopilot~\cite{fadadu2022multi} to virtual reality~\cite{ebner2017multi}. 
Although traditional MVS methods have achieved significant performance, many learning-based methods~\cite{yao2018mvsnet,surfacenet,gu2020cascade,unimvs,ding2022transmvsnet} have demonstrated their superior ability to tackle low-texture and repetitive pattern regions for a more accurate and complete reconstruction.

Generally, reconstructing a scene using learning-based multiview stereo (MVS) techniques involves two steps: depth prediction and fusion rendering. Learning-based methods primarily focus on optimizing the depth prediction process to provide accurate depth maps for subsequent fusion rendering. Therefore, the learning-based MVS reconstruction task can be seen as a depth prediction task. Recent works~\cite{ding2022transmvsnet,wang2022mvster,wei2021aa} have improved the accuracy of depth prediction by enhancing feature matching and cost regularization. Additionally, techniques such as deformable convolutions~\cite{wei2021aa,ding2022transmvsnet} and attention mechanisms~\cite{ding2022transmvsnet,wang2022mvster} have been utilized to obtain accurate depth maps. However, an interesting phenomenon has been observed: \textit{Depth maps with smaller estimation errors might not achieve better 3-D reconstruction quality after fusion rendering.}
Are there other factors limiting the accuracy of 3-D reconstruction? After a thorough investigation into the fusion process, we found that the depth geometry is an important factor that has been overlooked in MVS. 
Different depth geometries suffer from significant performance gaps, even for the same depth estimation error case. Thus it is worth considering \textit{what constitutes a good depth geometry}.

\begin{figure}[t]
  \centering
    \includegraphics[width=0.42\textwidth]{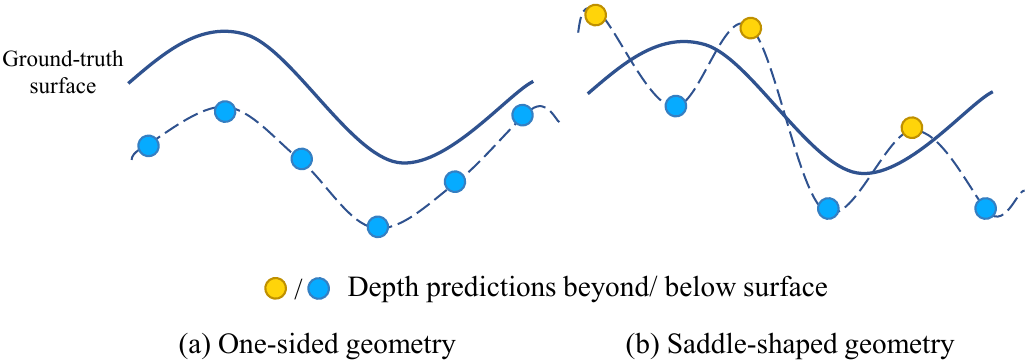}
  \caption{\textbf{Brief view of depth geometries.} There are 1-D views of the one-sided geometry and the saddle-shaped geometry.   } 
  \vspace{-0.2in}
  \label{fig:Brief view }
\end{figure}

To address the question, as shown in~\cref{fig:Brief view }, we introduced two ideal depth geometries representing two extreme cases: geometry composed of one-sided cells vs. geometry composed of a saddle-shaped cell. The former has depth planes on the same side as the ground-truth surface, while the latter oscillates back and forth on both sides of the ground-truth surface. To evaluate the impact of these depth geometries on 3D reconstruction, we artificially controlled cells of the predicted depth planes while ensuring the same absolute error in depth prediction. Interestingly, we found that saddle-shaped cells significantly improved 3D reconstruction performance compared to one-sided cells. We attribute this performance improvement to the depth interpolation operation during the fusion rendering stage, which is highly sensitive to the depth cell. Saddle-shaped cells can minimize the expected interpolation depth error and thus lead to enhanced 3D reconstruction performance. And we propose a novel method to obtain saddle-shaped depth cells.

The depth geometry with saddle-shaped cells requires the prediction of depth values in an oscillating pattern. However, due to the prior of surface continuity, neural networks tend to predict a smooth depth map for a scene. We propose a new method called Dual-Depth, which predicts two depth values for each pixel. To achieve an oscillating depth geometry, we first constrain the prediction error of the two depth values separately to ensure that they are both as close as possible to the ground-truth depth. Then, we jointly optimize these two predicted depths by constraining the interval of them. A novel checkerboard selection strategy is also proposed to combine the two depth values to obtain the final depth map. By embedding the above Dual-Depth method into a coarse-to-fine framework, we propose a novel MVS network, named Dual MVSNet (DMVSNet).

We conducted extensive experiments to demonstrate the suitability of the depth geometry with saddle-shaped cells for MVS and the effectiveness of our proposed method. Due to the Dual-Depth method, DMVSNet outperforms most methods in the indoor data set DTU and the outdoor data set (Intermediate Tanks and Temple). On the more challenging dataset (Advanced Tanks and Temple), DMVSNet achieved state-of-the-art performance, improving by $5.6\%$. These results highlight the importance of depth geometry for the MVS task and demonstrate that our proposed method is effective in improving the accuracy of 3D reconstructions. Moreover, our approach offers a new direction for future research, where the depth geometry can be exploited to enhance the reconstruction performance.

\vspace{5pt}
\noindent\textbf{Contribution.} 
In this work, we introduced a new perspective for considering depth geometry in MVS. We proposed the depth geometry with saddle-shaped cells for the first time and demonstrated its importance for the MVS reconstruction task. Technically, we proposed the dual-depth method to achieve saddle-shaped cells and designed the corresponding network framework DMVSNet. With the help of the Dual-Depth method, DMVSNet achieves top performance on the DTU dataset and state-of-the-art performance on the Tanks and Temple dataset.

\section{Relative Work}
\noindent\textbf{Traditional MVS.} There are four traditional MVS methods that can be classified based on their output: point-based~\cite{colmap,agarwal2011building}, volume pixels-based~\cite{seitz1997photorealistic}, mesh-based~\cite{fua1995object,wu2011high}, and depth maps-based~\cite{campbell2008using,galliani2015massively,schonberger2016pixelwise,colmap}. 
Among them, depth map-based methods break down the reconstruction task into two parts: depth prediction and fusion. Since depth prediction can be performed in parallel and requires only a subset of views, methods based on it are more flexible. After obtaining the estimated depth maps of all images, a fusion process is utilized to generate the 3-D point representation, which is the most commonly used in MVS methods and applications. 

\noindent\textbf{Learning-based MVS.} Despite traditional MVS methods demonstrating their advantages, they rely on the handcrafted similarity metric~\cite{unimvs}. In contrast, pioneering works such as surefacenet~\cite{surfacenet} and MVSNet~\cite{yao2018mvsnet} 
leverage the power of neural networks to generalize potential patterns and learn the metric from the data. 
MVSNet~\cite{yao2018mvsnet} introduces a learning depth map-based pipeline and is widely applied in the following works. R-MVSNet~\cite{yao2019recurrent} proposes recurrent structures on cost regularization for efficiency. The coarse-to-fine framework based on the pipeline is presented by~\cite{gu2020cascade,yang2020cost, cheng2020deep}.
The two essential components of depth prediction are feature matching and cost regularization. Improving the quality of feature representations can benefit both components, and recent research has focused on this aspect. Techniques such as deformable convolution~\cite{wei2021aa,ding2022transmvsnet} and attention mechanisms~\cite{ding2022transmvsnet,caomvsformer} have been used to obtain more precise depth maps, resulting in improved reconstruction quality.
However, accurate depth maps are not the only determining factor. In this paper, we will show that previously neglected depth geometry is critical as well.

\noindent\textbf{Depth prediction.} In addition to multiview depth prediction, there are two other types: Monocular depth prediction~\cite{roy2016monocular,wang2022less,eigen2014depth,xian2020structure} and stereo depth prediction~\cite{khamis2018stereonet,liang2018learning,chang2018pyramid}. The former is often used in visual effects that do not require a highly accurate depth map because of its inherently ill-posed nature. The latter is more accurate due to the epipolar constraint and can be applied in motion-sensing games and autonomous driving. However, for applications that require a precise and complete depth map, such as 3-D reconstruction, the shading issues inherent in stereo depth prediction can be mitigated by using multiview depth prediction. 
In the context of MVS, although many studies concentrate on the quality of depth maps, to our knowledge, there is no one that emphasizes the importance of the geometry of the estimated depth, which is the concern of this paper.

\section{Motivation}
\label{sec:motivation}
\begin{figure}[t]
  \centering
    \includegraphics[width=0.4\textwidth]{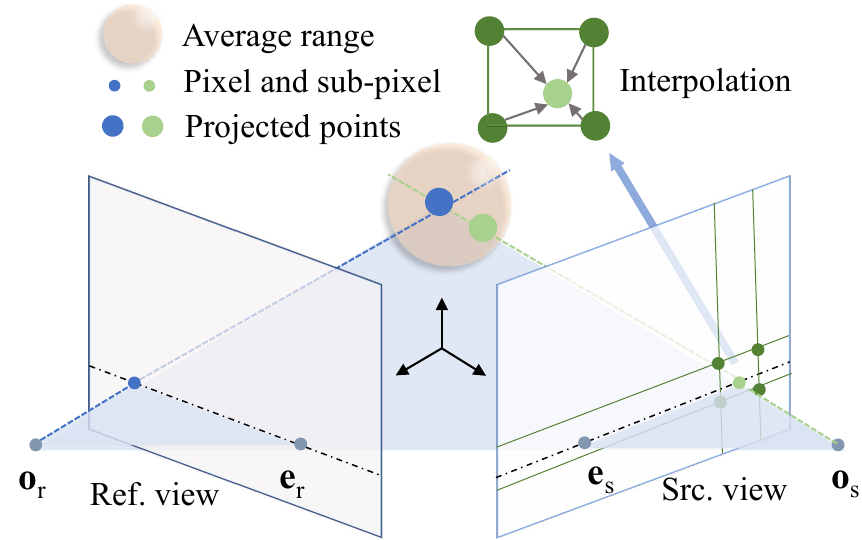}
  \caption{\textbf{Fusion process.} Projecting a pixel in the ref. view to the $3$-D space and further projecting it to the sub-pixels in the others views. The sub-pixels are reprojected to the $3$-D space with the interpolated depth of the depth estimations of their surrounding pixels. The final depth for the pixel in the ref. view to generate the $3$-D point is calculated by an averaging among the projected and reprojected points within a range around the projected point.} 
  \vspace{-0.1in}
  \label{fig:fusion process}
\end{figure}

\begin{table}[t]\Huge
  \centering
  \resizebox{1\linewidth}{!}{
  \begin{tabular}{@{}llll@{}}
    \toprule
    Settings & Acc.$\downarrow$ & Comp.$\downarrow$ & Overall$\downarrow$\\
    \midrule
    CasMVSNet               &0.366 &0.324 & 0.345  \\
    w. one-sided         &0.467~($-27.6\%$)&0.380~($-17.2\%$)&0.424~($-22.9\%$)\\
    w. saddle-shapped      &0.243~($+36.6\%$)&0.249~($+23.1\%$)&0.246~($+28.7\%$)\\
    \bottomrule
  \end{tabular}
}
  \vspace{-10pt}
\caption{{Results in DTU with different scenarios of~\cref{fig:different depth prediction}.} The depth error for them is the same of $10.47$mm.}
  \label{tab:motivation}
  \vspace{-8pt}
\end{table}

\subsection{Estimated bias and interpolated bias}
Given a reference image $\bm{I}_1 \in \mathbb{R}^{3 \times H \times W}$ and its source images $\{\bm{I}_i\}_{i=2}^{N}$, as well as their respective camera intrinsics and extrinsics estimated by image matching methods~\cite{superglue,zhao2023A2B,aaaiye, sun2021loftr,zhao2023learning}, MVS methods predict a depth map $\bm{D} \in \mathbb{R}^{H \times W}$ aligned with $\bm{I}_1$. The depth maps are then filtered and utilized to fuse $3$-D cloud points with the given camera intrinsics $\{\bm{K}_i\}_{i=1}^{N}$ and extrinsics $\{\bm{T}_i\}_{i=1}^{N}$. 

During the fusion process illustrated in Fig.~\ref{fig:fusion process}, the pixels in the reference view are projected onto a 3D point in space using the estimated depth map. This 3D point is then reprojected onto sub-pixels in other views using their respective camera parameters, and the corresponding depth maps are used to obtain new 3D points. The final 3D reconstruction result is determined by the depth differences of pixels in the reference view and the estimated depths of subpixels in other views (\textit{e.g.}, by averaging). Therefore, the accuracy of the 3D reconstruction result is affected not only by the accuracy of the estimated depth maps but also by the accuracy of the interpolated depths of subpixels. The subpixel depths are estimated by linearly interpolating the depths of neighboring pixels, and their accuracy is influenced by the estimation bias and depth cell\footnotemark[2]. Fig.~\ref{fig:SaddleLike_1d} shows that the accuracy of the interpolated depth can vary under the same estimation bias and interpolation position due to different depth cells. Therefore, it is important to consider the impact of the depth geometry with different cells for MVS.

\footnotetext[2]{The concept of ``Cell'' in this paper is similar to that in HOG~\cite{dalal2005histograms}.}

\begin{figure}[t]
  \centering
    \includegraphics[width=0.45\textwidth]{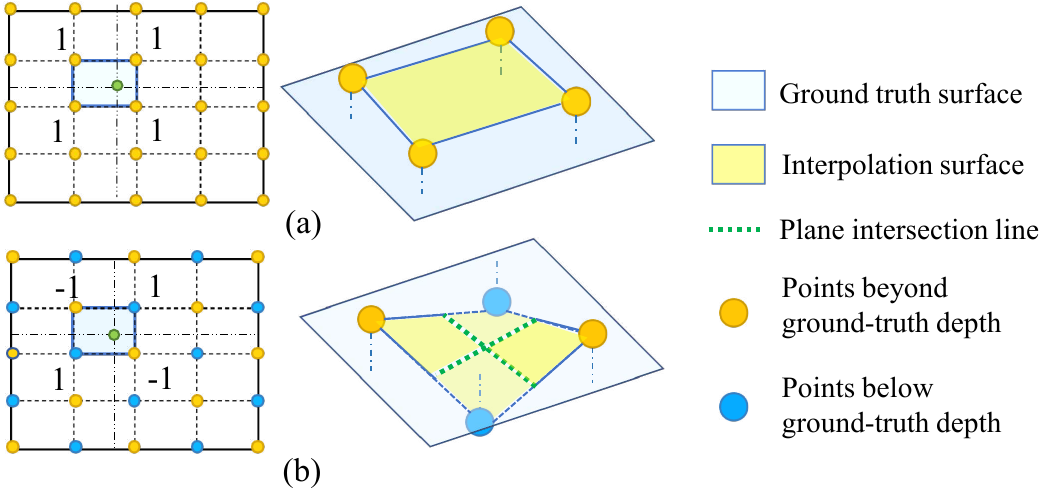}
  \caption{\textbf{Two kinds of depth cells with the same estimation bias.} (a)One-sided: all depths are on the same side~(beyond or below) of the ground truth surface and there is no intersection line between them. (b)Saddle-shaped: depths of two adjacent pixels are not on the same side of the ground truth and there are two plane intersection lines on any four adjacent pixels. The average absolute estimated bias of (a) and (b) are all `$1$'. The expectations of absolute interpolated bias are `$1$' and `$0.25$' respectively. } 
  \label{fig:different depth prediction}
\end{figure}

\begin{figure}[!t]
  \centering
    \includegraphics[width=0.45\textwidth]{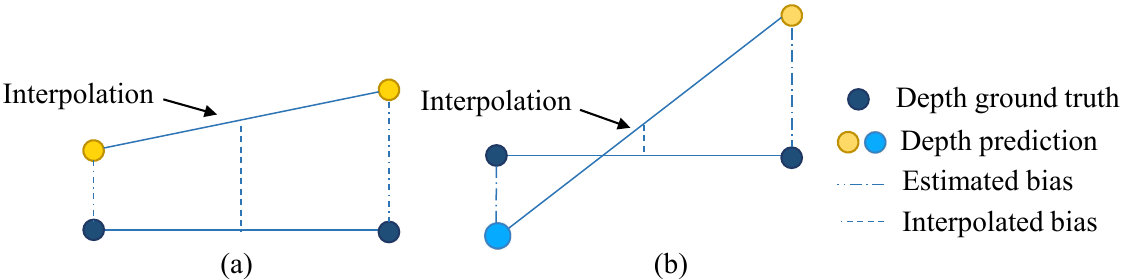}
  \caption{\textbf{Estimated bias and Interpolated bias.} Interpolated bias can have significant performance gaps because of the depth geometries (a) and (b), even under the same estimated bias.} 
  \vspace{-0.1in}
  \label{fig:SaddleLike_1d}
\end{figure}

\begin{table}[t]\small
  \centering
  \resizebox{0.8\linewidth}{!}{
  \begin{tabular}{@{}l|c|lll@{}}
    \toprule
    Settings & Depth Err.$\downarrow$ & ACC.$\downarrow$ & Comp.$\downarrow$ & Overall$\downarrow$\\
    \midrule
    +attention  &\textbf{9.15}&0.369&0.318&0.343\\
    +dcn       &9.76&\textbf{0.356}&\textbf{0.317}&\textbf{0.336}\\
    \bottomrule
  \end{tabular}
}
\caption{{A counter-intuitive phenomenon}. Despite better depth prediction performance, poorer 3D metrics were obtained.}
  \label{tab:mt}
  \vspace{-10pt}
\end{table}

\subsection{One-sided V.S. Saddle-shaped}

To briefly illustrate the difference of depth cells, we present two hypothetical depth cells in Fig.~\ref{fig:different depth prediction}:~a) One-sided cells;~b) Saddle-shaped cells. We assume that the interpolated positions with the same absolute estimation bias of `$1$' are uniformly distributed. The spatial volume between the depth plane (yellow) and the ground truth plane (blue) can be considered as the expected absolute interpolation error. Mathematically, the expected absolute interpolation error for the ``one-sided cell" is four times higher than that for the ``saddle-shaped cell".

To quantitatively demonstrate the impact of depth geometry with different cells on the performance of 3D point reconstruction, we conducted a toy verification experiment. Under the assumption that the absolute estimation bias of each pixel is the same, we flipped the estimated depth values using the true depth, making them distributed according to the two cells shown in Fig.~\ref{fig:different depth prediction}. The experimental results in Table~\ref{tab:motivation} show that depth geometries with different cells have a significant impact on the quality of 3D point reconstruction, including accuracy and completeness, even with differences in accuracy exceeding $60\%$ (the second and third rows in Table~\ref{tab:motivation}). This indicates that the depth geometry with saddle-shaped cells is a feasible approach to improve the performance of MVS.

Most existing MVS methods do not impose constraints on depth cells, so their depth maps are distributed between geometries composed of one-sided and saddle-shaped cells, which determines the accuracy of 3D point reconstruction that falls between the performance of the two ideal geometries composed of singular cells (as shown in the first row of Table~\ref{tab:motivation}). 
Besides, without constraints on depth cells, despite better depth prediction performance, poorer 3D metrics may be obtained(Table~\ref{tab:mt}).
How can we constrain the network to generate a depth map with more saddle-shaped cells? In the above toy experiment, the method of flipping the estimated depth using the ground truth is a chicken-and-egg problem, which is not feasible in practical inference. In the next section, we will introduce a dual-depth prediction to address this dilemma.

\begin{figure}[t]
  \centering
    \includegraphics[width=0.4\textwidth]{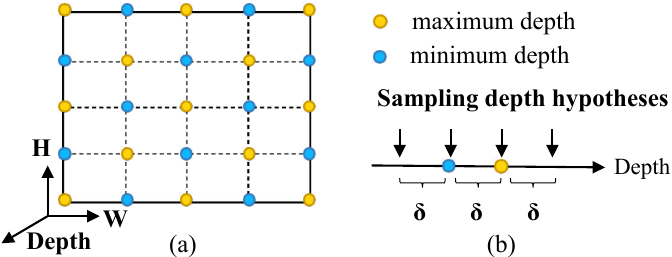}
  \caption{\textbf{Checkerboard selection strategy and depth hypotheses sampling.} (a) We alternately select the minimum or maximum depth of the two predictions for each pixel. (b) We define the euclidean distance between the two estimated depths corresponding to a pixel as its uncertainty, which serves as the interval for the next round of depth sampling.} 
  \label{fig:coordinate show}
  \vspace{-10pt}
\end{figure}

\section{Dual-Depth Prediction}
\subsection{Review of learning-based MVS}
\label{sec:review}

In traditional learning-based MVS pipelines, a weight-shared CNN is used to first extract feature maps $\{\bm{F}_i\in\mathbb{R}^{F^{'}\times H^{'}\times W^{'}}\}_{i=1}^{N}$ aligned with images $\{\bm{I}_i\}_{i=1}^{N}$, where $H^{'}$, $W^{'}$, and $F^{'}$ represent the height, width, and number of channels of the feature map, respectively.  
The depth hypotheses for a pixel $\{d_i\}_{i=1}^{M}$ is usually uniformly sampled within the range of $[\alpha_1,\alpha_2]$.
With the depth hypotheses, as well as camera intrinsic matrix $\bm{K}$ and extrinsic matrix $\bm{T}$, a differentiable homography transformation is used to construct a feature volume  $\{\bm{V}_i\in\mathbb{R}^{F^{'}\times M\times H^{'} \times W^{'}}\}_{i=1}^{N}$ in $3$D space. At depth $d$, the homography matrix between the $k$-th view and the reference camera frustum is given by:

\begin{equation}
    \bm{H}_k^{d}=d\bm{K}_k\bm{T}_k\bm{T}_1^{-1}\bm{K}_1^{-1}\,.
    \label{eq:homography matrix}
\end{equation}
For the pixels $\bm{p}\in \mathbb{R}^{2 \times H^{'} \times W^{'}}$ in the reference image, the transformed pixels in the image of the $k$-th view at depth $d$ are
\begin{equation}
    \bm{p}_{k}^{d}=\bm{H}_k^{d}\bm{p}_{1}\,.
    \label{eq:transformed pixels}
\end{equation}
The feature volumes are constructed by warping feature maps from source images to the reference camera frustum per to pixels and their transformed pixels at different depth hypotheses per to~\cref{eq:transformed pixels}. By regularizing the cost volume generated by measuring the similarity of the feature volumes, a probability volume $\bm{P}\in\mathbb{R}^{M\times H^{'} \times W^{'}}$ can be obtained. The depth of a pixel located at coordinates $(x,y)$ in the reference view can be obtained by using the following.
\begin{equation}
    \overline{\bm{D}}{(x,y)}=\sum_{i=1}^{M}d_i\bm{P}{(i,x,y)}\,.
    \label{eq:generate depth raw}
\end{equation}

\noindent\textbf{Loss function.} 
In common MVS methods, $L_1$ loss is used to supervise the estimated depth map $\overline{\bm{D}}$ with
\begin{equation}
    {L}_{est}(\overline{\bm{D}},\bm{D}_g)=L_1(\overline{\bm{D}},\bm{D}_g)\,,
    \label{eq:common loss}
\end{equation}
where $\bm{D}_{g}$ is the ground-truth depth map. $L_{est}$ aims to minimize the difference between the estimated depth map and the ground-truth depth map, thus reducing the estimated bias. However, it lacks the ability to enforce the geometry of the estimated depth, let alone predict a saddle-shaped depth map. Furthermore, the objective of the saddle-shaped cell depth map is inconsistent with the objective of $L_{est}$, which encourages the estimated depth map to approach the smooth depth map of the truth of the ground.

\subsection{Dual-Depth}
\label{sec:dual depth}

  Aiming at an oscillating depth geometry  with more saddle-shaped cells, we choose to predict two depth values for each pixel. If the dual depth is distributed on either side of the ground-truth depth, a heuristic selection strategy can achieve the target geometry.

 \begin{figure}[t]
  \centering
    \includegraphics[width=0.45\textwidth]{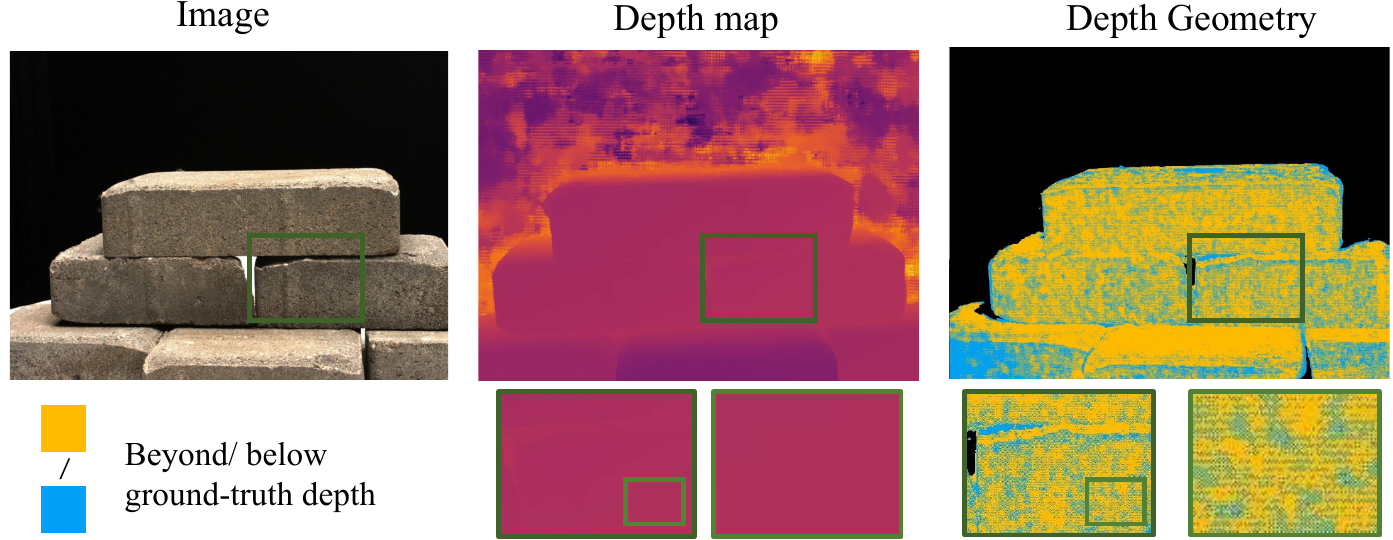}
  \caption{\textbf{Visualization of the depth geometry with saddle-shaped cells.} We colored the pixels whose estimated depth value is beyond/below the ground truth with orange/blue. 
  } 
  \label{fig:beyond below view}
  \vspace{-10pt}
\end{figure}

Specifically, we generate two probability distributions for each pixel and use them to generate two corresponding depth maps $\bm{D}\in \mathbb{R}^{2 \times H^{'} \times W^{'}}$ (see Eq. \eqref{eq:generate depth raw}).
To ensure the accuracy of the independently predicted dual depth, we take a $L_1$ loss to supervise their predicted values, as in previous works. Intuitively, without adding constraints on the joint distribution of the dual depth, the resulting prediction distribution is disordered. Therefore, we propose another novel loss to constrain the two depths to be symmetrically distributed around the ground truth.
\begin{equation}\small
 \begin{split}
    {L}_{int}(\bm{D},\bm{D}_g)&=L_1(|{\tt max}(\bm{D})-{\tt min}(\bm{D})|,\\
    &{max}(|{\tt max}(\bm{D})-\bm{D}_g|,|{\tt min}(\bm{D})-\bm{D}_g|))\,,
 \end{split}
    \label{eq:loss 2}
\end{equation}
 where $|\cdot|$ indicates the absolute distance, ${\tt max}(\cdot)$ and ${\tt min}(\cdot)$ takes the maximum and minimum value along the first dimension, \textit{e.g.} ${\tt max}(\bm{D})= max(\bm{D}[1, :, :], \bm{D}[2, :, :])$. $L_{int}$ encourages estimated bias is no larger than $|{\tt max}(\bm{D})-{\tt min}(\bm{D})|$, such that the interval increases as estimated bias increasing, which guarantees that dual-depth is distributed on either side of the ground-truth depth. If ${\tt max}(\bm{D})={\tt min}(\bm{D})=\bm{D}_g$, $L_{int}$ reaches the minimal value, suggesting an unbiased depth estimate and is consistent with the objective of $L_{est}$.

When the true depth value lies between the predicted dual-depth, We propose a checkerboard selection strategy to choose the appropriate depth prediction value for each pixel. Specifically, we alternate between selecting the maximum and minimum predicted depth values, creating a distribution that resembles a checkerboard. As shown in Fig.~\ref{fig:coordinate show}(a), the depth of pixel $(x,y)$ is determined by
 \begin{equation} \small
    \bm{D}_{c}(x,y)=\begin{cases} {\tt min}(\bm{D})(x,y)\,, &{x\%2 == y\%2}\\
                                {\tt max}(\bm{D})(x,y)\,, &otherwise\end{cases}
                                \,,
    \label{eq: depth 1}
\end{equation}
which generates an oscillating depth map $\bm{D}_{c}$\,.
As shown in Fig.~\ref{fig:beyond below view}, the depth map obtained by the dual-depth method achieves the geometry composed of saddle-shaped cells. The depth map within the box is smooth, indicating that the predicted values are close to the ground truth. At the same time, its corresponding depth geometry presents a saddle-shaped form, which is consistent with our expectations.

However, the above approach might carry a potential risk of increasing depth prediction errors when the true depth value at $(x,y)$ is not within the range of ${\tt min}(\bm{D})(x,y)$ and ${\tt max}(\bm{D})(x,y)$. To address this issue, we propose using Cascade Dual-Depths, which will be illustrated in the next section.

\begin{figure}[!t]
\centering
    \includegraphics[width=0.45\textwidth]{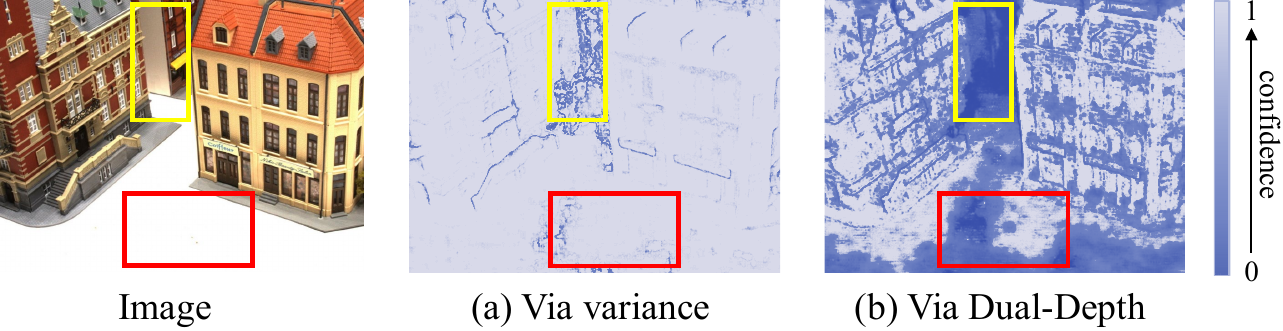}
  \caption{\textbf{Comparison of confidence maps}. 
  The confidence map via Dual-Depth provides more accurate confidence for occluded regions or invalid backgrounds.
  The lighter region indicates more higher confidence. 
  }
  \vspace{-0.3in}
  \label{fig:confidence map}
\end{figure}

\begin{figure*}[t]
  \centering
    \includegraphics[width=0.95\textwidth]{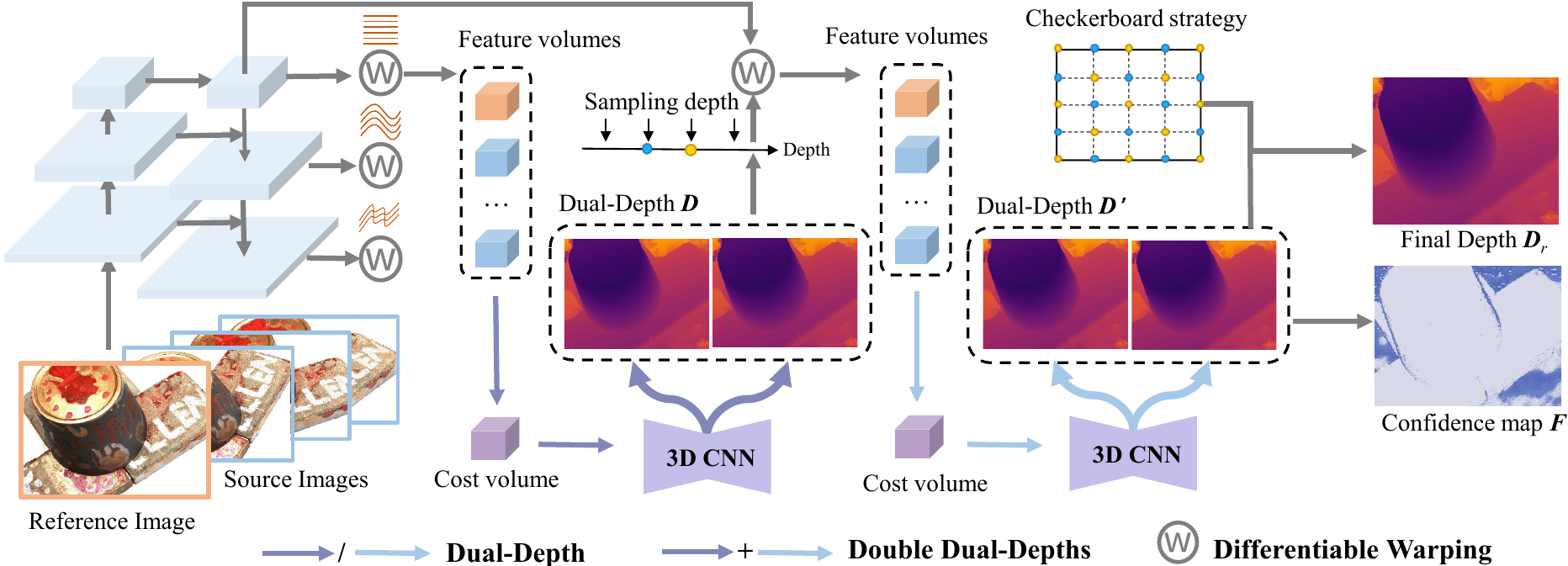}
  \caption{\textbf{Architecture details of DMVSNet.} Our backbone adopts a classical coarse-to-fine framework.
  We employ a shared feature pyramid network for extracting feature maps, and then warp them to obtain a cost volume. Next,  we use a 3D convolutional neural network (CNN) to generate two depth maps, denoted as Dual-Depth $\bm{D}$. By resetting the depth sampling range using $\bm{D}$, we repeat the depth prediction process to obtain another dual-depth $\bm{D}^{'}$, which is utilized to construct the final depth $\bm{D}_r$ with the checkerboard strategy. Per to Eq.~\eqref{eq:confidence map}, the corresponding confidence map $\bm{F}$ can be calculated.
  } 
  \label{fig:DMVSNet}
  \vspace{-5pt}
\end{figure*}

\subsection{Cascade Dual-Depths}
\label{sec:ddd}
Despite the fact that the encouraging estimated bias is not larger than $|{\tt max}(\bm{D})-{\tt min}(\bm{D})|$ in the dual depth, the uncovered issue occurs when the estimated bias is too large, which is beyond the range of $|{\tt max}(\bm{D})-{\tt min}(\bm{D})|$. The reason is that the fixed range of depth hypotheses $\alpha_2 - \alpha_1$ leads to a large depth estimation bias. Intuitively, when a pixel's estimated depth is unreliable, the range of depth hypotheses should be increased to ensure that the ground truth is included in the searching space. In contrast, the range can be appropriately narrowed for more reliable estimates. For example, UCS-Net~\cite{cheng2020deep} leverages the variance of the probability distribution to reflect uncertainty and dynamically adjusts the range of depth hypotheses for the corresponding sampling depths, resulting in smaller estimation biases. Inspired by it, we attempted to utilize uncertainty estimates to adaptively adjust the search range of depth hypotheses. 

We first adopt the variance of the probability distribution to obtain a confidence map, similar to UCS-Net (as shown in Fig.~\ref{fig:confidence map}(a)). 
However, the confidence map obtained through variance tends to predict similar confidence levels for most regions, making it unreliable for predicting confidence levels in weakly textured areas or at edges.
Additionally, there exists inference conflict between the two corresponding confidence values since each pixel predicts the dual-depth. Therefore, we need to find an appropriate way to represent the uncertainty of the dual-depth, not solely relying on the probability distribution.

In daily life, when measuring an object with a ruler, people usually take two measurements and compare the results. If there is a large difference between the two measurements, the precision of the measurement is considered low. Similarly in dual-depth estimation, if the difference between the maximum and minimum depth values predicted by a pixel is large, we consider the estimation bias to be large. Consequently, the depth searching space for this pixel will be enlarged in the next iteration. To adaptively adjust the range of the depth hypotheses, we use the absolute distance between ${\tt max}(\bm{D})(x,y)$ and ${\tt min}(\bm{D})(x,y)$ as the boundary of the depth hypotheses, as shown in \cref{fig:coordinate show}(b). Given the range of depth hypotheses, we can construct feature volumes, cost volumes, and probability distributions as described in \cref{sec:review}. Following the principles outlined in \cref{sec:dual depth}, we calculate the refined dual depth $\bm{D}'$, which is then used to obtain the final depth $\bm{D}_r$ according to \cref{eq: depth 1}.

\noindent{\textbf{Confidence Map.}} 
The confidence map is used in the fusion process to mask out pixels with substantially deviated depths using a threshold, thereby preventing their projection into the 3D space. In this paper, the confidence is given by
\begin{equation} \label{eq:confidence map}
    \bm{F}(x,y)=2{\tt sigmoid}(\frac{1}{\bm{U}(x,y)})-1 \in \left( 0,1 \right)\,,
\end{equation}
where $\bm{U}(x,y)=|{\tt max}(\bm{D})(x,y) - {\tt min}(\bm{D})(x,y)|$. The confidence map is shown in \cref{fig:confidence map}(b).

\begin{table}\Huge
  \centering
  \resizebox{1\linewidth}{!}{
  \begin{tabular}{@{}lcccc@{}}
    \toprule
    Method &Years& ACC.(mm)$\downarrow$ & Comp.(mm)$\downarrow$ & Overall(mm)$\downarrow$\\
    \midrule
    Gipuma~\cite{gipuma} &ICCV2015&\textbf{0.283} &0.873 &0.578\\
    COLMAP~\cite{colmap} &CVPR2016&0.400 &0.664 &0.532\\
    SurfaceNet~\cite{surfacenet}&ICCV2017 &0.450 &1.040 &0.745\\
    MVSNet~\cite{yao2018mvsnet} &ECCV2018&0.396 &0.527 &0.462 \\
    P-MVSNet~\cite{luo2019p}&ICCV2019 &0.406 &0.434 &0.420\\
    R-MVSNet~\cite{yao2019recurrent}&CVPR2019 &0.383 &0.452 &0.417\\
    Point-MVSNet~\cite{chen2019point}&ICCV2019 &0.342 &0.411 &0.376\\
    CasMVSNet~\cite{gu2020cascade}&CVPR2020 &0.325 &0.385 &0.355\\
    CVP-MVSNet~\cite{yang2020cost}&CVPR2020 &\underline{0.296} &0.406 &0.351 \\
    UCS-Net~\cite{cheng2020deep} &CVPR2020&0.338 &0.349 &0.344\\
    AA-RMVSNet~\cite{wei2021aa} &ICCV2021&0.376 &0.339 &0.357\\
    UniMVSNet~\cite{unimvs} &CVPR2022&0.352 &0.278 &0.315\\
    transMVSNet~\cite{ding2022transmvsnet}&CVPR2022 & 0.321 &0.289 &\textbf{0.305} \\
    MVSter~\cite{wang2022mvster}&ECCV2022&0.350&\underline{0.276}&\underline{0.313}\\
    
    \hline
    DMVSNet & - &0.349 &\underline{0.276} &\underline{0.313} \\
    DMVSNet$^{\star}$ & - &0.338 &\textbf{0.272} &\textbf{0.305} \\
    \bottomrule
  \end{tabular}
  }
  \vspace{-2pt}
  \caption{{Results on DTU.} We report our results with a vanilla checking strategy and dynamic
checking strategy$^{\star}$. The best performance is in boldface and the second best is underlined.).
  }
  \label{tab:dtu}
  \vspace{-10pt}
\end{table}

\begin{table*}\huge
  \centering
  \resizebox{1\linewidth}{!}{
  \begin{tabular}{@{}l|c|ccccccccc|ccccccc@{}}
    \toprule
    \multirow{2}{*}{Method}&\multirow{2}{*}{Years} & \multicolumn{9}{c|}{Intermediate($\%$)$\uparrow$} & \multicolumn{7}{c}{Advanced($\%$)$\uparrow$} \\ 
                             && Mean& Fam.& Fra.&Hor. &Lig. &M60. &Pan. &Pla. &Tra. & Mean& Aud.&Bal. &Cou. &Mus.&Pal. & Tem.\\
    \midrule
    PointMVS~\cite{chen2019point}&ICCV2019                &48.27&61.79&41.15&34.20&50.79&51.97&50.85&52.38&43.06&   - &   - &  -  &  -  & -  & -   & -   \\
    PatchmatchNet~\cite{wang2021patchmatchnet}   &CVPR2021        &53.15&66.99&52.64&43.24&54.87&52.87&49.54&54.21&50.81&32.31&23.69&37.03&30.04&41.80&28.31&32.29\\
    CVP-MVSNet~\cite{yang2020cost}&CVPR2020       &54.03&76.50&47.74&36.34&55.12&57.28&54.28&57.43&47.54&   - &   - &  -  &  -  & -  & -   & -   \\
    CasMVSNet~\cite{gu2020cascade}         & CVPR2020     &56.84&76.37&58.45&46.26&55.81&56.11&54.06&58.18&49.51&31.12&19.81&38.46&29.10&43.87&27.36&28.11\\
    UCS-Net~\cite{cheng2020deep} &CVPR2020      &54.83&76.09&53.16&43.03&54.00&55.60&51.49&57.38&47.89&   - &   - &  -  &  -  & -  & -   & -   \\
    D2HC-RMVSNet~\cite{yan2020dense} &ECCV2020      &59.20&74.69&56.04&49.42&60.08&59.81&59.61&60.04&53.92&   - &   - &  -  &  -  & -  & -   & -   \\
    AA-RMVSNet~\cite{wei2021aa} & ICCV2021    &61.51&77.77&59.53&51.53&\textbf{64.02}&{64.05}&59.47&60.85&55.50&33.53&20.96&40.15&32.05&46.01&29.28&32.71\\
    EPP-MVSNet~\cite{ma2021epp}        &ICCV2021     &61.68&77.86&60.54&52.96&62.33&61.69&60.34&\textbf{62.44}&55.30&35.72&21.28&39.74&35.34&49.21&30.00&38.75\\
    UniMVSNet~\cite{unimvs} &  CVPR2022  &\underline{64.36}&\underline{81.20}&\underline{66.43}&53.11&\underline{63.46}&\textbf{66.09}&\textbf{64.84}&\underline{62.23}&\underline{57.53}&\underline{38.96}&\underline{28.33}&44.36&\underline{39.74}&\underline{52.89}&33.80&34.63\\
    transMVSNet~\cite{ding2022transmvsnet}&  CVPR2022  &63.52&{80.92}&65.83&\underline{56.94}&62.54&63.06&60.00&60.20&\textbf{58.67}&37.00&24.84&\underline{44.59}&34.77&46.49&\underline{34.69}&36.62\\
    MVSter~\cite{wang2022mvster} &ECCV2022&60.92&80.2&63.51&52.30	&61.38&	61.47&	58.16	&58.98&	51.38 &37.53&26.68&	42.14&	35.65&	49.37&	32.16&	\underline{39.19}\\
    \midrule
    Ours& -  & \textbf{64.66} &\textbf{ 81.27}   & \textbf{67.54}  &  \textbf{59.10} & 63.12  & \underline{64.64} &\underline{64.80} &  59.83 & 56.97 & \textbf{41.17} & \textbf{30.08 }  &\textbf{46.10}   &\textbf{40.65 }  &\textbf{53.53}   &\textbf{35.08 }& \textbf{41.60} \\
    
    \bottomrule
  \end{tabular}
  }
  \vspace{-2pt}
  \caption{{Quantitative results on Tanks and Temples benchmark.} We report the F-score metric and ``Mean'' refers to the average F-score of all scenes. The best performance is in \textbf{boldface} and the second best is \underline{underlined}.}
  \label{tab:temple and tank}
  \vspace{-10pt}
\end{table*}
\subsection{DMVSNet}
\label{sec:dmvsnet}
To embed our double-dual depths method into the multi-view stereo (MVS) task, we propose a coarse-to-fine MVS framework, named DMVSNet. As shown in Fig.~\ref{fig:DMVSNet}, the dual depths structure is incorporated into the depth regression stage through differentiable warping.

Specifically, we adopt a Feature Pyramid Network~(FPN)~\cite{lin2017feature} to extract multiscale features like CasMVSNet~\cite{gu2020cascade}, and double the output channels to obtain feature maps for a cascading dual-depth. Then, we construct feature volumes by warping feature maps via uniformly sampled depth hypotheses. The feature volumes are then aggregated into cost volumes using a similarity metric, such as the inner-product-based metric. A $3$-D CNN is utilized to transform the cost volume into the probability distribution. To obtain dual-depth $\bm{D}$, we double the 3-D CNN to obtain two probability distributions and generate depth maps using~\cref{eq:generate depth raw}. We repeat this process with adaptively sampled depth hypotheses to generate refined dual-depth $\bm{D}^{'}$, which is utilized to construct the final depth $\bm{D}_r$ with the checkerboard selection strategy.

 \vspace{5pt}
\noindent{\textbf{Training Loss.}}
We adopt $L_{est}$ and $L_{int}$ to respectively reduce estimation bias and interpolation bias. 
Since the interpolated biases result from the depth error of sub-pixels, we additonally supervise the depths of sub-pixels at coordinates $(x+0.5, y+0.5)$.
The final loss function is 
\begin{equation}\footnotesize
    \begin{split}
        &L=L_{est}(\bm{D},\bm{D}_g)+L_{int}(\bm{D},\bm{D}_g)+L_{sub}(\bm{D},\bm{D}_g)\\
        &+L_{est}(\bm{D}^{'},\bm{D}_g)+L_{int}(\bm{D}^{'},\bm{D}_g)+L_{sub}(\bm{D^{'}},\bm{D}_g)
    \end{split}\,,
    \label{eq:tot loss}
\end{equation}
where $L_{sub}(\bm{D}),\bm{D}_g)=L_1({\tt sub.}({\tt con.}(\bm{D}),\bm{D}_g))$, ${\tt con.}(x)$ suggests constructing a depth map with the checkerboard selection strategy,and ${\tt sub.}$ indicates taking the sub-pixels at coordinates $(x+0.5, y+0.5)$.


\section{Experiments}

\noindent{\textbf{Datasets.}}
We conducted training and evaluation of our models using the DTU dataset~\cite{dtu}. We fine-tune the BlendedMVS~\cite{yao2020blendedmvs} and subsequently evaluate on Tanks and Temples~\cite{tanks} for generalization. The DTU dataset comprises 124 indoor scenes that were captured under controlled camera and lighting conditions. To ensure consistency, We adapt the same training and evaluation split as MVSNet~\cite{yao2018mvsnet}. On the other hand, the BlendedMVS dataset is a synthetic dataset with 113 scenes that simulate both indoor and outdoor conditions. We follow the UniMVS and adapt the same training and validation split. Lastly, the Tanks and Temples benchmark comprises scenes captured in a complex and realistic environment and serves as an online benchmark. It is divided into intermediate and advanced sets based on the level of difficulty.

\noindent{\textbf{Metrics.}} 
For the depth metric, we utilize the absolute distance of the disparity between the predicted depth and the ground truth depth, commonly referred to as depth error or estimated bias. As for the 3-D representation, we report the standard metrics, namely accuracy, completeness, and overall score, using the official evaluation toolkit.

\noindent{\textbf{Implementation Details.}} 
We performed network optimization over 16 epochs using a learning rate of 0.001, with semi-decay occurring in epochs 10, 12, and 14. During the evaluation process, we used the final depth $\bm{D}_r$ to calculate the metrics and perform the fusion. To ensure consistency with DTU standards, the input images were resized to $1152\times864$, and the number of input images was set to $5$. Before evaluating on Tanks and Temples, our network underwent a fine-tuning process on BlendedMVS for $10$ epochs, following established protocols. For training and evaluation, the number of input images was set to $9$ and $11$, respectively. Consistent with previous research~\cite{yan2020dense}, we also adopted the dynamic check strategy for fusion.

\subsection{Results on DTU}
\label{sec:dtu}
    To assess the effectiveness of our proposed approach, we adopt a vanilla checking strategy. Drawing inspiration from MVSNet~\cite{yao2018mvsnet}, we use confidence maps and geometric constraints for depth filtering. Specifically, we set the probability threshold and the minimum number of consistent views to $0.3$ and $5$, respectively. Our baseline comprises traditional and learning-based MVS methods. 
    Compared with previously published works, our vanilla check strategy achieves the highest completeness and the second-highest overall results, as depicted in Table \ref{tab:dtu}.
    It is important to note that the fusion strategies used in previous works are not standardized, such as transMVSNet~\cite{ding2022transmvsnet} used the dynamic check strategy~\cite{yan2020dense}. For a fair comparison, we also present our method result with the dynamic check strategy$^{\star}$, which outperforms all other methods in terms of the overall performance.

\subsection{Results on Tanks and Temples}
Following the common setting, we evaluate the efficacy of our method in terms of generalization using the Tanks and Temples online benchmark, after fine-tuning on BlendedMVS. As shown in~\cref{tab:temple and tank}, we report the F-score that is defined as a harmonic mean of accuracy and completeness. Our method achieves SOTA performance in both intermediate and advanced sets. In the advanced set, a more challenging subset of Tanks and Temples, we achieve the state-of-the-art results in every scene, with an $5.6\%$ improvement over the second-best result. 
It is worth noting that the accuracy of depth estimation for a scene is inversely proportional to the difficulty of that scene. Dual-depth prediction may not mitigate the bias in estimated depth, but it reduces interpolated bias via the depth geometry, indicating that it better fits for scenes with inaccurate depth estimation.

\begin{figure*}[t]
  \centering
    \includegraphics[width=0.95\textwidth]{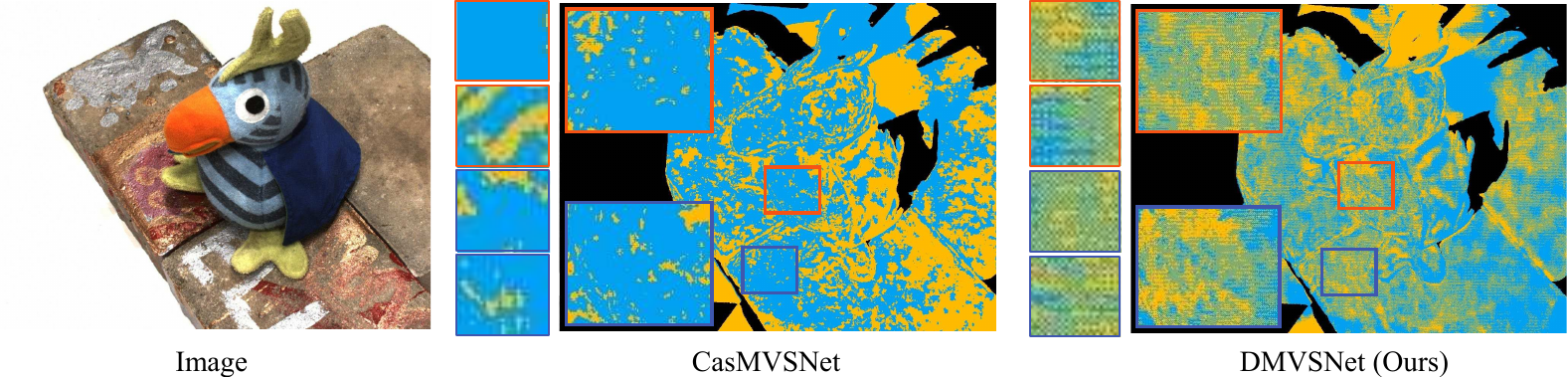}
  \caption{\textbf{The comparison of the depth geometries.} We visualize the depth geometry by coloring the pixel whose estimated depth is beyond the ground truth with orange and the others with blue.} 
  \label{fig:understand}
  \vspace{-10pt}
\end{figure*}

\section{Ablation Study}
%


    \vspace{5pt}
    \noindent\textbf{Which part works.} We first report the results of the ablation study in~\cref{tab:decompose}. 
    The baseline choose the CasMVSNet framework.
    By introducing Dual-Depth, we observe a significant $7.5\%$ enhancement in the $3$-D representation. The Cascade Dual-Depths bring another $1.6\%$ improvement over Dual-Depth in $3$-D representation.  
    It should be noted that the advancement in depth error~(\textit{i.e.}, the estimated bias), is not the primary factor that contributes to the improved 3-D representation achieved by Dual-Depth. Instead, the saddle-shaped geometry plays a crucial role. 
    
    To further illustrate the importance of the checkerboard strategy, we conduct an experiment in~\cref{tab:selecting strategies}.
    We fixed the parameters of the model and only changed the depth selection strategy. The results showed that while the checkerboard strategy had little effect on the accuracy of depth prediction, it significantly improved 3-D reconstruction metrics. When the checkerboard strategy was not used, i.e., only one side with two predicted depth values was selected (such as selecting the minimum depth), the depth values of adjacent pixels were closer due to the lack of constraints. This made adjacent pixels more likely to be on the same side of the true surface, which disrupted some saddle-shaped cells that should have existed in the depth map using the checkerboard strategy. Therefore, the checkerboard strategy is indispensable for saddle-shaped depth geometry.
\begin{table}\large
  \centering
    \resizebox{0.9\linewidth}{!}{
  \begin{tabular}{@{}l|c|ccc@{}}
    \toprule
    Settings & Depth Err. & ACC. & Comp. & Overall\\
    \midrule
    w.o. Dual-Depth      &10.47 &0.366 &0.324 & 0.345  \\
    w.o. Cascade Dual-Depths&10.03 &0.352&0.288&0.320\\
    DMVSNet~(Ours) &\textbf{9.12} &\textbf{0.349}&\textbf{0.276}&\textbf{0.313}\\
    \bottomrule
  \end{tabular}
  }
  \vspace{+2pt}
  \caption{ {Ablation studies on DTU.}  We apply the same fusion setting as in~\cref{sec:dtu}
  }
  \label{tab:decompose}
\end{table}

\begin{figure}[t]
  \centering
    \includegraphics[width=0.45\textwidth]{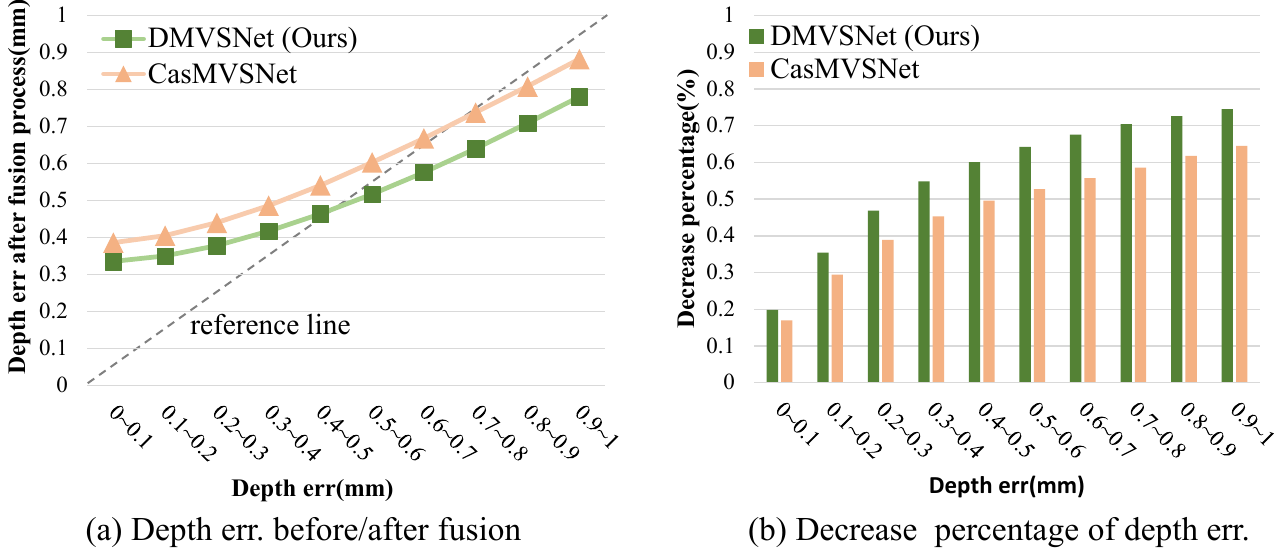}
  \caption{\textbf{How Dual-Depth works.} Compared with side prediction, the estimated biases of dual-depth decrease more(a). The ratios of depths whose estimated bias decreases are shown in (b) and ratios of Dual-Depth are higher in all intervals.  
  } 
  \label{fig:explanation}
    \vspace{-5pt}
\end{figure}

    \vspace{5pt}
    \noindent\textbf{What Dual-Depth does.} 
    To understand the importance of Dual-Depth, we conducted an interesting experiment in scene 29 of DTU, as shown in~\cref{fig:explanation}. We analyzed the pixel depth error changes after fusion within the range of $0\sim1$ mm. A reference line is characterized by zero-change depth error. 
    As shown in~\cref{fig:explanation}(a), our DMVSNet achieves smaller depth errors after fusion compared with CasMVSNet, indicating that the Cascade Dual-Depths can help reduce initial depth errors during fusion. As shown in Fig. \ref{fig:explanation}(b), we illustrate the percentage of pixels with reduced errors after fusion for each depth error interval. Our method achieves a higher proportion of depth error reduction in all error intervals, which explains our superior performance in Fig.~\ref{fig:explanation}(a).
    We display the geometry of Dual-Depth Prediction and the baseline in~\cref{fig:understand}, to understand what Dual-Depth Prediction does. The Dual-Depth Prediction presents depth predictions at the pixel level in an oscillating pattern, while the baseline method tends to predict depth regionally smooth. Specifically, the Dual-Depth Prediction generates a checkerboard-like depth geometry that approximates the expected geometry composed of saddle-shaped cells.

  \vspace{5pt}
    \noindent\textbf{Why is Dual-Depth. needed.} 
    To better understand the contribution of the method details, we conducted additional ablation studies in Table \cref{tab:replace}. Firstly, we removed the constraints on the geometric shape of the depth estimation (by not using $L_{int}$ and $L_{sub}$), which resulted in decreased depth and 3-Dreconstruction quality. Then, we did not apply the double branches in the 3-D CNN. Although it had a similar depth estimation accuracy, the 3-D reconstruction results were worse. These observations demonstrate the necessity of constraints on depth geometry and additional structural design.

\begin{table}\Huge
  \centering
  \resizebox{0.95\linewidth}{!}{
  \begin{tabular}{@{}l|c|ccc@{}}
    \toprule
    Settings & Depth Err. & ACC. & Comp. & Overall\\
    \midrule
    
    w.o. checkerboard strategy     &10.05 &0.386&0.321&0.354\\ 
    w. checkerboard strategy~(Ours)    &\textbf{10.03}&\textbf{0.352}&\textbf{0.288}&\textbf{0.320}  \\ 
    \bottomrule
  \end{tabular}
  }
  \vspace{+2pt}
  \caption{{Evaluations w.o./w. checkerboard strategy.} We take the fixed model to obtain depth maps w.o./w. checkerboard strategy. With similar qualities of depth estimation, the checkerboard strategy achieves a significant improvement in the performance of 3D reconstruction.}
  \label{tab:selecting strategies}
\end{table}

\begin{table}\huge
  \centering
  \resizebox{0.82\linewidth}{!}{
  \begin{tabular}{@{}l|c|ccc@{}}
    \toprule
     Settings & Depth Err. & ACC. & Comp. & Overall\\
    \midrule
    w.o. $L_{int}$ and $L_{sub}$ &11.09 &0.356&0.318 &0.337 \\
    w.o. double branches &9.26  &0.356	&0.297	&0.327 \\
    Cascade Dual-Depths (Ours)  &\textbf{9.12} &\textbf{0.349}&\textbf{0.276}&\textbf{0.313}\\
    \bottomrule
  \end{tabular}
  }
  \vspace{+2pt}
  \caption{{Additional experiments on loss functions and network architecture.}}
  \label{tab:replace}
  \vspace{-5pt}
\end{table}
\section{Conclusion}
In this article, we present a novel perspective to enhance the performance of learning-based MVS by incorporating constraints of the depth geometry. We demonstrate that the proposed saddle-shaped cells outperform other cells in terms of qualitative and quantitative measures. We introduce the Dual-Depth, checkerboard selection strategy, and Cascade Dual-Depths to implement the saddle-shaped depth geometry. By integrating them into a coarse-to-fine framework, we develop a novel DMVSNet approach. Extensive experiments prove that DMVSNet achieves superior performance in 3-D reconstruction and the importance of depth geometry.

\vspace{+5pt}
\noindent \textbf{Acknowledgments.}
This work was supported in part by the National Natural Science Foundation of China (Grant No. U1913602) and funded by the Huawei Technologies CO., LTD.

{\small
\bibliographystyle{ieee_fullname}
\bibliography{egbib}
}

\clearpage
\appendix
\begin{appendices}
\section{Mathematic Expectations of\\ Interpolated Biases}
\label{sec: expectations}
In the motivation section, we present the effect of depth structures both qualitatively and quantitatively. Here, we report the details about mathematic expectations of different interpolated biases.

\vspace{10pt}
\noindent{\textbf{Bilinear interpolation.}} Bilinear interpolation is considered as a multiplexing of the linear interpolation. Due to the facts that the interpolated result is affected by surrounding four pixels and the pixel grid can be viewed as $1$, the bilinear interpolation of the estimated bias map $\triangle\bm{D}$ can be expressed as 
\begin{equation}\small
\begin{split}
    \triangle\bm{D}(x,y)&=\triangle\bm{D}(\lfloor x\rfloor ,\lfloor y\rfloor )*(\lfloor x+1\rfloor -x)*(\lfloor y+1\rfloor -y)\\
                &+\triangle\bm{D}(\lfloor x+1\rfloor ,\lfloor y\rfloor )*(x-\lfloor x\rfloor )*(\lfloor y+1\rfloor -y)\\
                &+\triangle\bm{D}(\lfloor x\rfloor ,\lfloor y+1\rfloor )*(\lfloor x+1\rfloor -x)*(y-\lfloor y\rfloor )\\
                &+\triangle\bm{D}(\lfloor x+1\rfloor ,\lfloor y+1\rfloor )*(x-\lfloor x\rfloor )*(y-\lfloor y\rfloor )
\end{split}\,,
\label{eq:interpolation}
\end{equation}
where $\lfloor x \rfloor$ indicates the function $\tt{floor(x)}$. The expectation of absolute interpolated biases can be calculated as
\begin{equation}
    \int_{\lfloor x \rfloor}^{\lfloor x+1 \rfloor} \int_{\lfloor y \rfloor}^{\lfloor y+1 \rfloor} |\triangle\bm{D}(x,y)|*\frac{1}{S}dydx\,,
    \label{eq:expectation}
\end{equation}
where $|\cdot|$ denotes the absolute operarion and $S=(\lfloor x+1 \rfloor-\lfloor x \rfloor)*(\lfloor y+1 \rfloor-\lfloor y \rfloor)=1$.

\vspace{10pt}
\noindent{\textbf{Expectation of the one-sided cell.}} 
One-sided cell indicates all depths are on the same side (beyond or below) of the ground truth.
To briefly express the expectation, we assume the estimated bias are all ``1'', which suggests $\triangle\bm{D}(\lfloor x\rfloor ,\lfloor y\rfloor )=\triangle\bm{D}(\lfloor x+1\rfloor ,\lfloor y\rfloor )=\triangle\bm{D}(\lfloor x\rfloor ,\lfloor y+1\rfloor )=\triangle\bm{D}(\lfloor x+1\rfloor ,\lfloor y+1\rfloor )=1$. Bringing them into~\cref{eq:interpolation,eq:expectation}.The expectation of the one-sided cell is
\begin{equation}
    \int_{\lfloor x \rfloor}^{\lfloor x+1 \rfloor} \int_{\lfloor y \rfloor}^{\lfloor y+1 \rfloor} |1|dydx=1\,.
    \label{eq:si}
\end{equation}

\vspace{10pt}
\noindent{\textbf{Expectation of the saddle-shaped cell.}} 
Saddle-shaped cell suggests depths of two adjacent pixels are not on the same side of the ground truth.
To briefly express the expectation, we assume $\triangle\bm{D}(\lfloor x\rfloor ,\lfloor y\rfloor )=\triangle\bm{D}(\lfloor x+1\rfloor ,\lfloor y+1\rfloor )=1$ and $\triangle\bm{D}(\lfloor x+1\rfloor ,\lfloor y\rfloor )=\triangle\bm{D}(\lfloor x\rfloor ,\lfloor y+1\rfloor )=-1$. The expectation of the saddle-shaped cell is 
\begin{equation}
\begin{split}
    \int_{\lfloor x \rfloor}^{\lfloor x+1 \rfloor} \int_{\lfloor y \rfloor}^{\lfloor y+1 \rfloor}&\\
    &|(\lfloor x+1\rfloor -x)*(\lfloor y+1\rfloor -y)\\
    &-(x-\lfloor x\rfloor )*(\lfloor y+1\rfloor -y)\\
    &-(\lfloor x+1\rfloor -x)*(y-\lfloor y\rfloor )\\
    &+(x-\lfloor x\rfloor )*(y-\lfloor y\rfloor )|dydx\\
    =0.25\,.
\end{split}
\label{eq:sa}
\end{equation}

According to the results in~\cref{eq:sa,eq:si}, the mathematic expectation of absolute interpolation error for the ``one-sided cell" is four times larger than that for the``saddle-shaped cell".

\section{Qualitative Results of Our Saddle-shaped Depth Geometry}
As a supplementary of Fig.~9 in the main text, we display more visualizations of the depth geometries of our estimated depth maps in~\cref{fig:depth geometry}. The Dual-Depth Prediction presents depth predictions at the pixel level in an oscillating pattern, such that there are many saddle-shaped cells in the depth maps. As discussed in main text and~\cref{sec: expectations}, the saddle-shaped cells are more helpful to decrease the interpolated bias under the same level of depth estimated quality and contribute to a better quality of 3-D reconstruction. The depth geometry with more saddle-shaped cells is the main factor that DMVSNet achieve top results in both indoor and outdoor datasets.

\section{Reliable Confidence Maps}
\begin{figure}[t]
  \centering
    \includegraphics[width=0.45\textwidth]{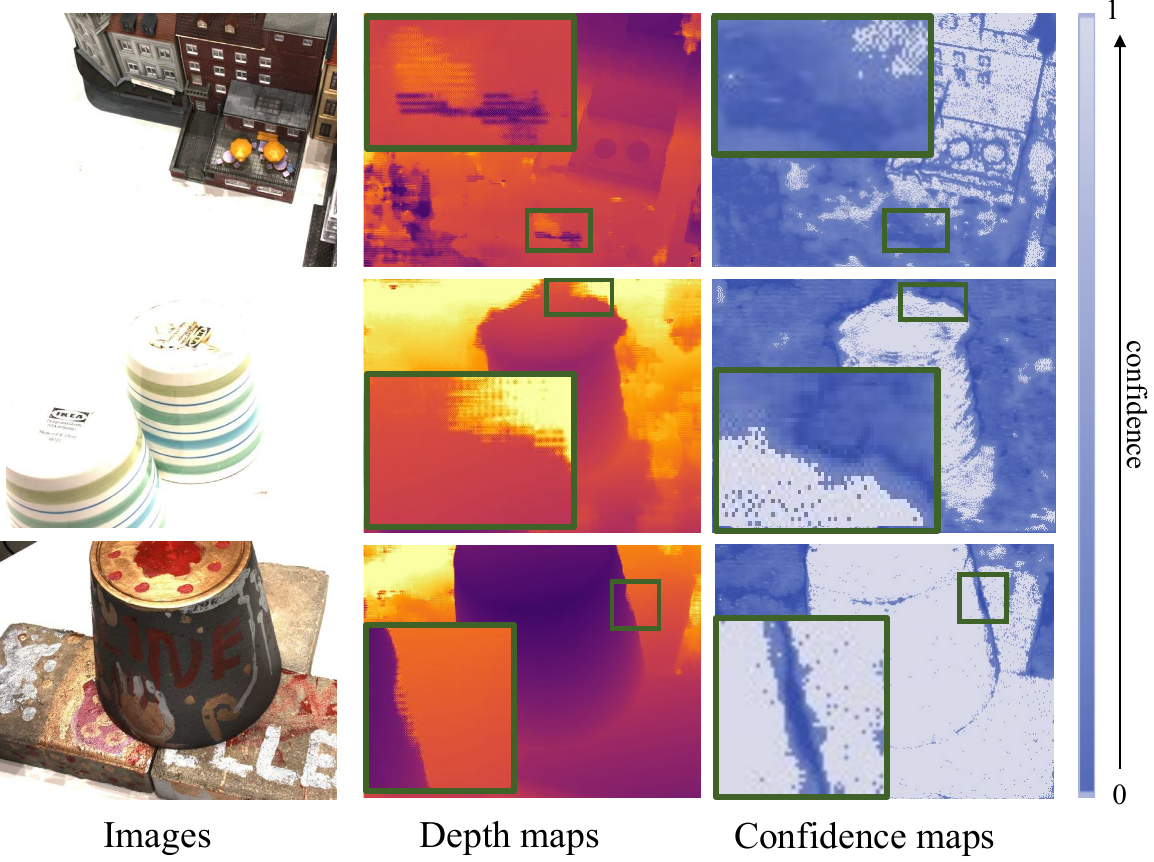}
  \caption{\textbf{Visualization of depths and their confidence maps.} We list some images with weak-texture, reflective, and edge regions and visualize their depth maps and confidence maps. } 
  \label{fig:s condifence}
\end{figure}

\begin{figure*}[t]
  \centering
    \includegraphics[width=0.85\textwidth]{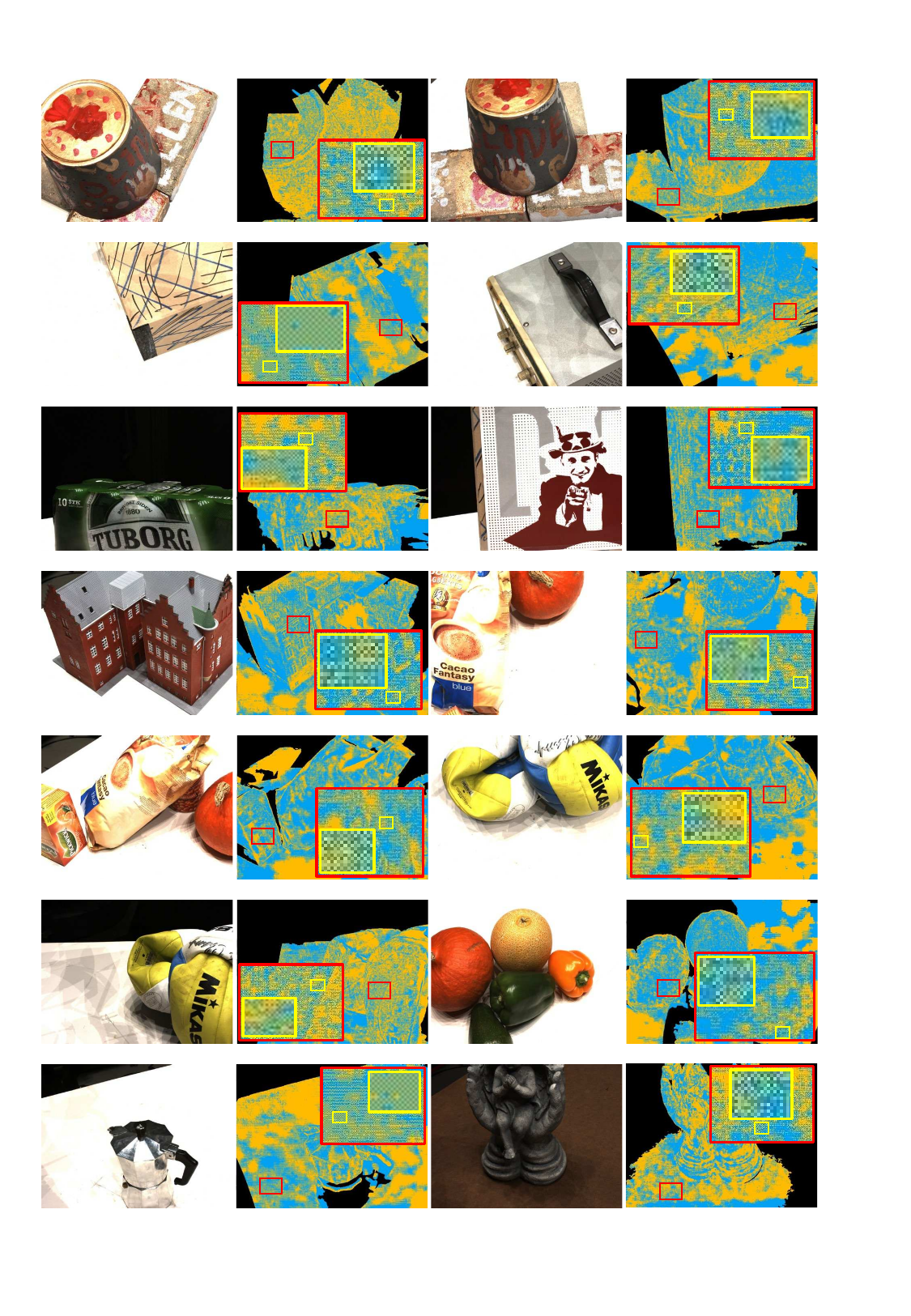}
  \caption{\textbf{More visualizations of our depth geometry.} We visualize the depth geometry by coloring the pixel whose estimated depth is beyond the ground truth with orange and the others with blue.} 
  \label{fig:depth geometry}
\end{figure*}

In the main text, we demonstrate the superiority of the confidence map defined with the interval of two depths against that of variance. We additionally show the paired depth map and confidence map in~\cref{fig:s condifence}. Generally, it is more difficult for network to predict accurate depths for  weak-texture, reflective, and edge regions. 
DMVSNet predicts an greatly oscillating pattern on these regions, results in depths with higher uncertainties. 
However, thanks to the confidence maps generated by Dual-Depth, the confidence level for these high-uncertainty regions is much lower, such that we can distinguish them. 

\section{Inference Time}
The structural designed for saddle-shaped cells introduces additional parameters because of the doubled 3-D CNNs, which would cost more inference time. As shown in Line $2\sim 3$ of~\cref{tab:inference times}, the Dual-Depth and Cascade Dual-Depths process more parameters and computational overheads. Since the Dual-Depth 
 requires double branches of 3-D CNNs to produce two depths for each pixel, the increasing cost is inevitable. Desipte more 
  cost, considering the improvements in the quality of 3-D reconstructions, our structural design for saddle-shaped is necessary and effective.

\begin{table}\Huge
  \centering
    \resizebox{0.95\linewidth}{!}{
  \begin{tabular}{@{}l|ccc|c@{}}
    \toprule
    Settings & Param.~(M) & Macs~(G) & Time~(ms) & Overall\\
    \midrule
    w.o. Dual-Depth      &0.94 &268 &220 & 0.345  \\
    w.o. Cascade Dual-Depths&1.80 &394&373&0.320\\
    DMVSNet & 2.67 &502&424&0.313\\ 
    \bottomrule
  \end{tabular}
  }
  \vspace{+2pt}

   \caption{ {Inference time, parameters, and reconstruction quality.} }
  \label{tab:inference times}
\end{table}

\section{More Qualitative Results}
\begin{figure*}[t]
  \centering
    \includegraphics[width=0.95\textwidth]{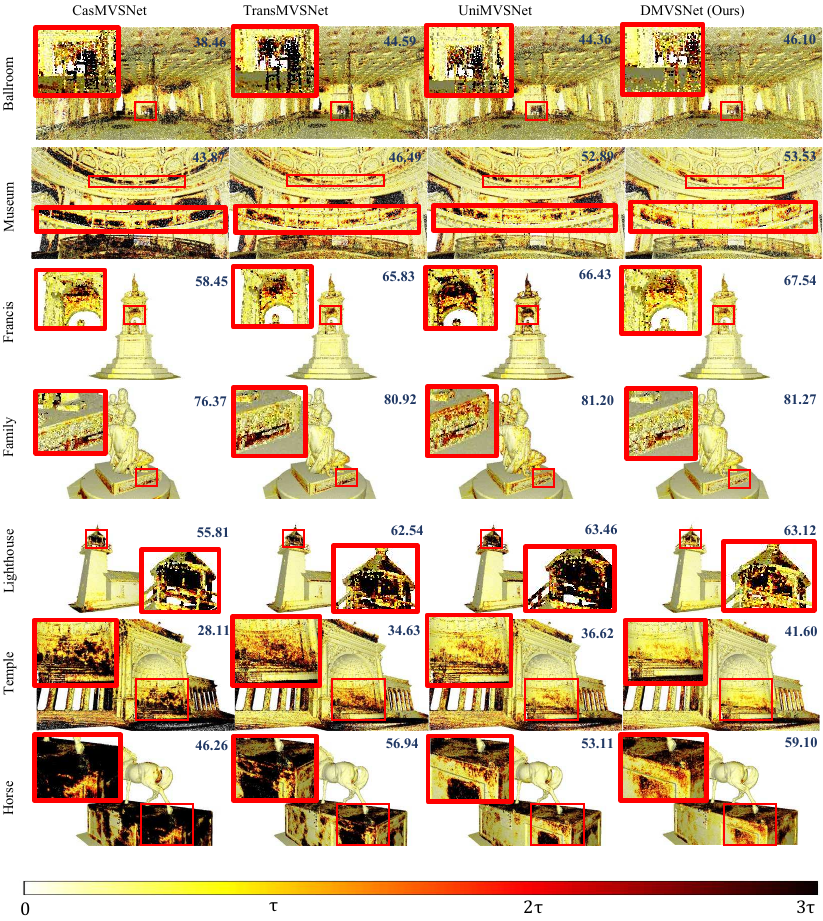}
  \caption{\textbf{Qualitative comparison on Tanks-and-Temples.} We visualize the completeness error of some sc as well as their F-scores. The $\tau$s of them are 10mm, 10mm, 5mm, 3mm, 5mm, 15mm, and 3mm respectively.  } 
  \label{fig: Tanks comparison}
\end{figure*}

\begin{figure*}[t]
  \centering
    \includegraphics[width=0.95\textwidth]{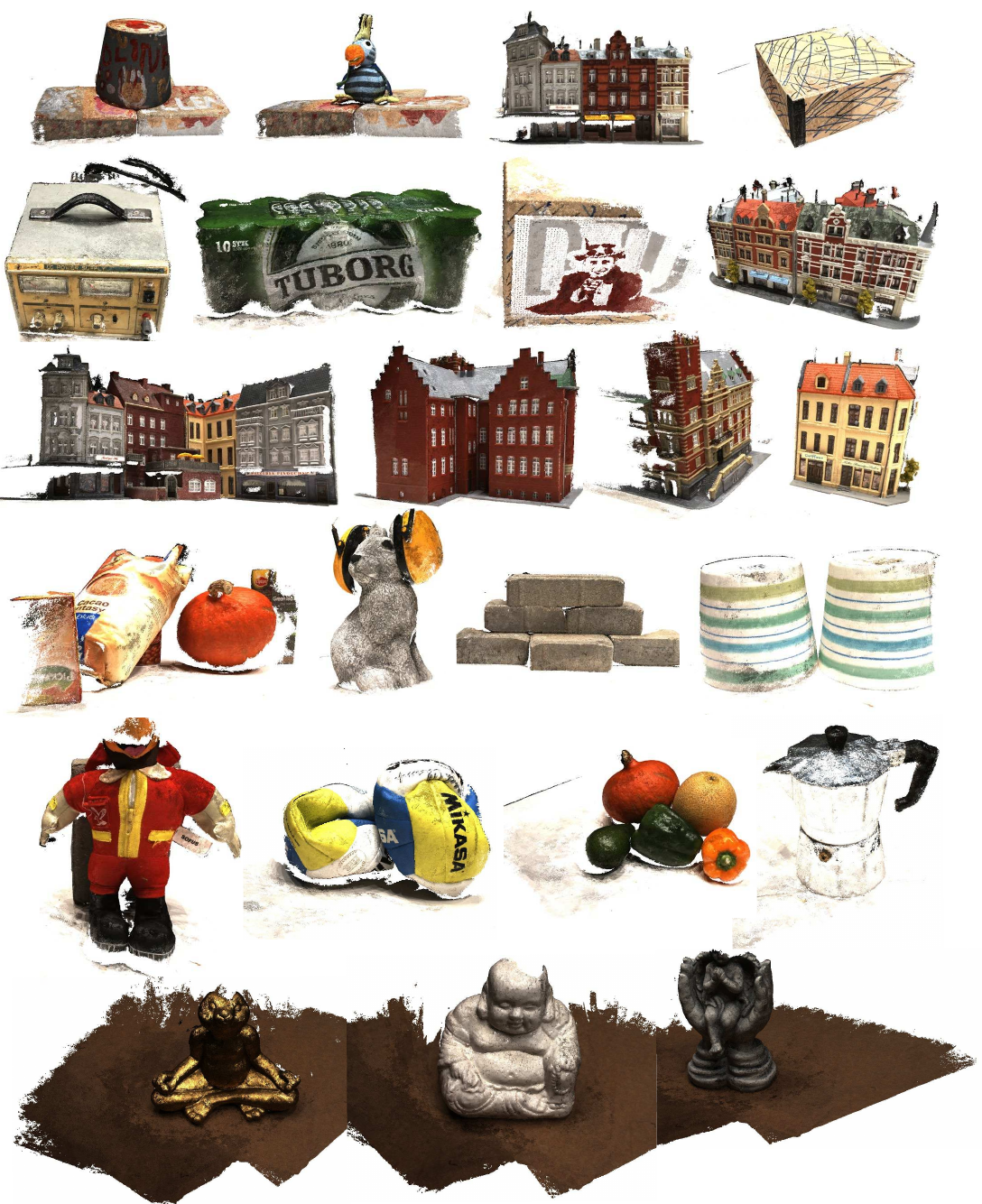}
  \caption{\textbf{3-D Point clouds.} We visualize our 3-D Point clouds of scenes in DTU evaluation set.  } 
  \label{fig:DTUPoints}
\end{figure*}

\begin{figure*}[t]
  \centering
    \includegraphics[width=0.90\textwidth]{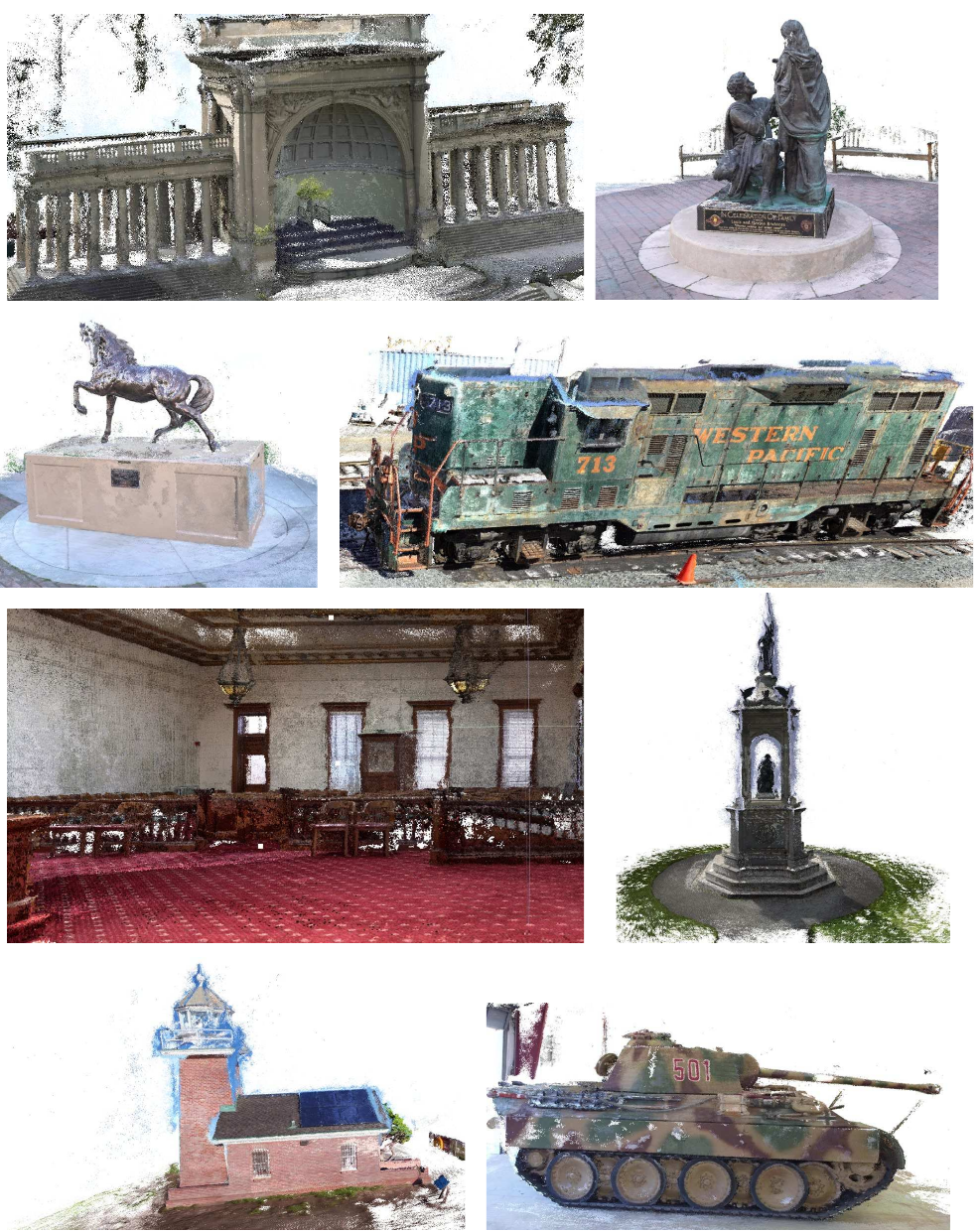}
  \caption{\textbf{3-D Point clouds.} We visualize our 3-D Point clouds of scenes in Tanks-and-Temples.  } 
  \label{fig:TanksPoints}
\end{figure*}

We visualize more qualitative results of DTU~\cite{dtu} and Tanks-and-Temples~\cite{tanks} benchmarks. In~\cref{fig:DTUPoints,fig:TanksPoints}, the 3-D point reconstructions of our approach are both dense and accurate in indoor and outdoor scenes, which demonstrates the robustness and scalability of DMVSNet on well-controlled scenarios and complex reality scenarios. Besides, we visualize the completeness error maps of scenarios in Tanks-and-Temples and report their F-scores on the top-right of visualizations in~\cref{fig: Tanks comparison}. Compared with CasMVSNet~\cite{gu2020cascade},TransMVSNet~\cite{ding2022transmvsnet}, and UniMVSNet~\cite{unimvs}, our 3-D point reconstructions are more complete as there are fewer dark red areas in the completeness error maps.

\end{appendices}

\end{document}